\documentclass{article}

\usepackage[nonatbib,preprint]{neurips_2019}

\usepackage[utf8]{inputenc} 
\usepackage[T1]{fontenc}    
\usepackage{hyperref}       
\usepackage{url}            
\usepackage{booktabs}       
\usepackage{amsfonts}       
\usepackage{nicefrac}       
\usepackage{tabularx}
\usepackage{microtype}      
\usepackage{amsmath}
\usepackage{graphicx}
\usepackage{subcaption}
\usepackage{xspace}
\usepackage{ifthen}
\newcommand{\sub}[1]{{_{\mathrm{#1}}}}

\newcommand{\figref}[1]{Fig. \ref{#1}}

\newcommand{\ExampleWidth}{0.32}
\newcommand{\ExampleRow}[9]{
    \begin{subfigure}[b]{\ExampleWidth\textwidth}
        \includegraphics[width=\textwidth]{#3}
        \ifthenelse{\equal{#9}{1}}{\caption{}}{}
    \end{subfigure}
    \begin{subfigure}[b]{\ExampleWidth\textwidth}
        \includegraphics[width=\textwidth]{#5}
        \ifthenelse{\equal{#9}{1}}{\caption{}}{}
    \end{subfigure}
    \begin{subfigure}[b]{\ExampleWidth\textwidth}
        \includegraphics[width=\textwidth]{#7}
        \ifthenelse{\equal{#9}{1}}{\caption{}}{}
    \end{subfigure}
      }

\newcommand{\ExamplesWidth}{0.245}
\newcommand{\Examples}[9]{
    \begin{subfigure}[b]{\ExamplesWidth\textwidth}
        \includegraphics[width=\textwidth]{#1}
        \ifthenelse{\equal{#9}{1}}{\caption{}}{}
    \end{subfigure}
    \begin{subfigure}[b]{\ExamplesWidth\textwidth}
        \includegraphics[width=\textwidth]{#2}
        \ifthenelse{\equal{#9}{1}}{\caption{}}{}
    \end{subfigure}
    \begin{subfigure}[b]{\ExamplesWidth\textwidth}
        \includegraphics[width=\textwidth]{#3}
        \ifthenelse{\equal{#9}{1}}{\caption{}}{}
    \end{subfigure}
    \begin{subfigure}[b]{\ExamplesWidth\textwidth}
        \includegraphics[width=\textwidth]{#4}
        \ifthenelse{\equal{#9}{1}}{\caption{}}{}
    \end{subfigure}\\
    \begin{subfigure}[b]{0.3\textwidth}
        \includegraphics[width=\textwidth]{#5}
        \ifthenelse{\equal{#9}{1}}{\caption{}}{}
    \end{subfigure}
    \begin{subfigure}[b]{0.3\textwidth}
        \includegraphics[width=\textwidth]{#6}
        \ifthenelse{\equal{#9}{1}}{\caption{}}{}
    \end{subfigure}
    \begin{subfigure}[b]{0.3\textwidth}
        \includegraphics[width=\textwidth]{#7}
        \ifthenelse{\equal{#9}{1}}{\caption{}}{}
    \end{subfigure}
      }

\newcommand{\DatasetExamples}[2]{

\begin{figure}
    \centering 			   
    \Examples{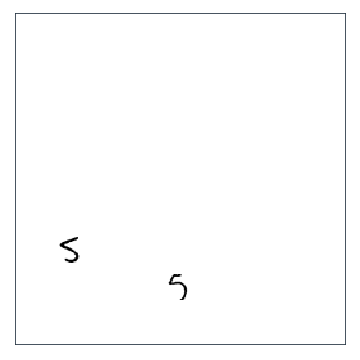}
			   {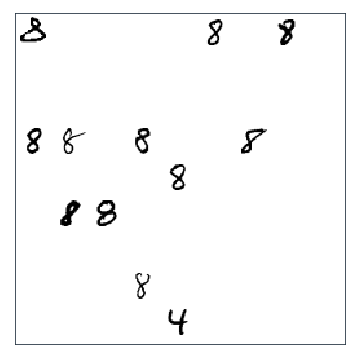}
			   {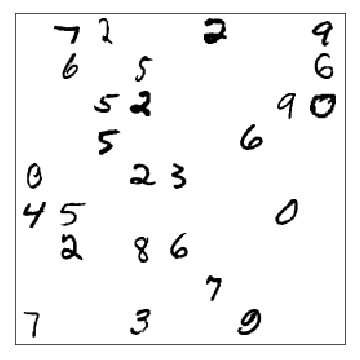}
			   {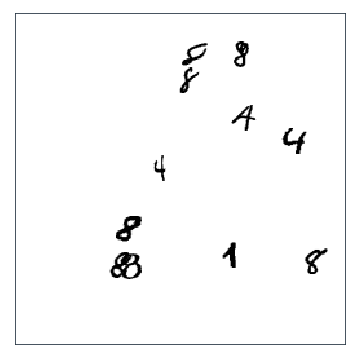}
			   {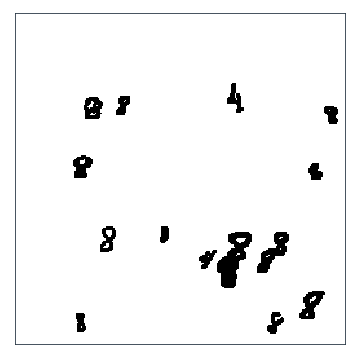}
			   {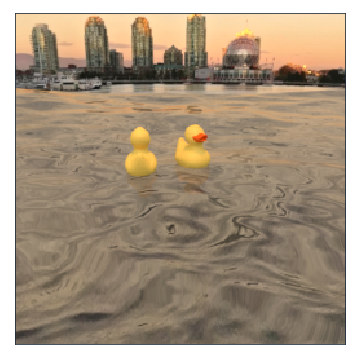}
			   {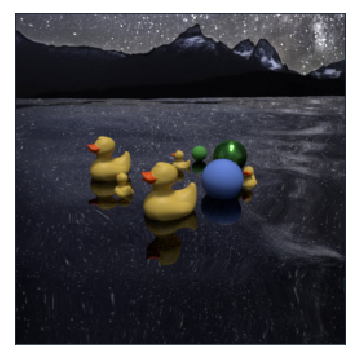}
			   {plotsrd-2_f8c12_resultspngexample0006.png}
			   {1}	

    \caption{#1}\label{#2}
\end{figure}

}

\newcommand{\ResultExamples}[2]{

\begin{figure}
    \centering
    \ExampleRow{plots/mnist-1_f8c8_results/png/density_maps/0_0.png}
			   {plots/mnist-2_f8c8_results/png/density_maps/0_0.png}
			   {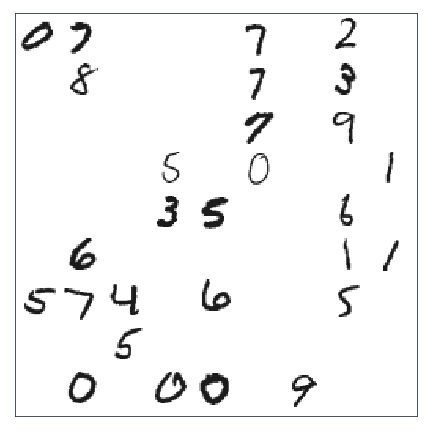}
			   {plots/mnist-2-occ_f8c8_results/png/density_maps/0_0.png}
			   {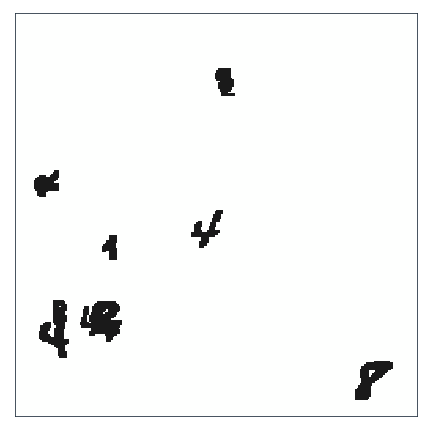}
			   {plots/rd-1_f8c12_results/png/density_maps/0_0.png}
			   {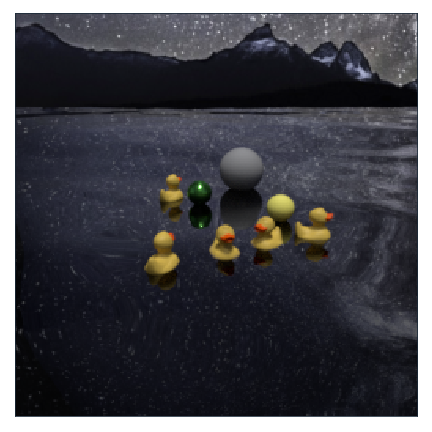}
			   {plotsrd-2_f8c12_resultspngdensity_maps2_0.png}
			   {0}	
    \ExampleRow{plots/mnist-1_f8c8_results/png/density_maps/0_1.png}
			   {plots/mnist-2_f8c8_results/png/density_maps/0_1.png}
			   {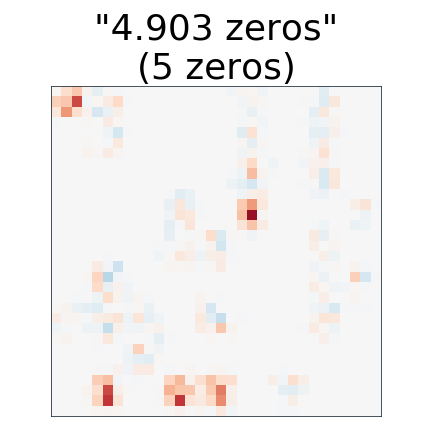}
			   {plots/mnist-2-occ_f8c8_results/png/density_maps/0_1.png}
			   {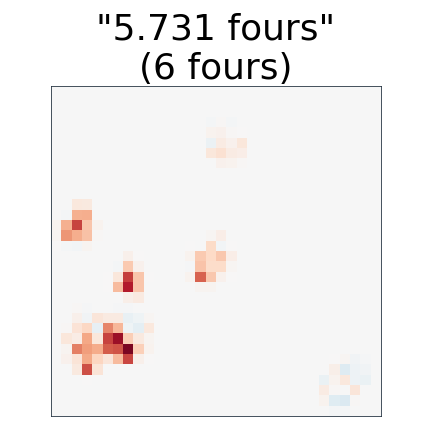}
			   {plots/rd-1_f8c12_results/png/density_maps/0_1.png}
			   {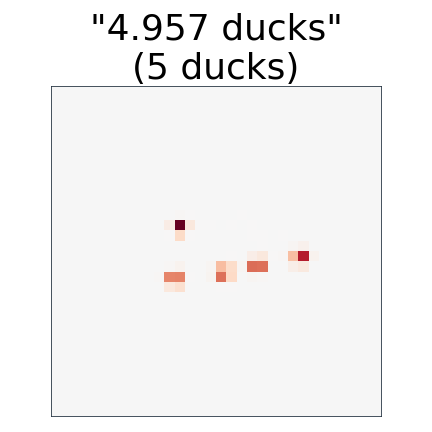}
			   {plotsrd-2_f8c12_resultspngdensity_maps2_1.png}
			   {0}	
	\ExampleRow{white.png}
			   {plots/mnist-2_f8c8_results/png/density_maps/0_2.png}
			   {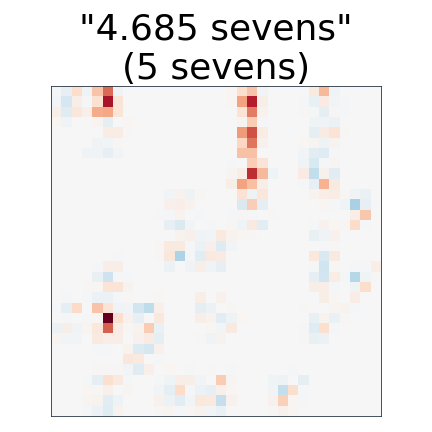}
			   {plots/mnist-2-occ_f8c8_results/png/density_maps/0_2.png}
			   {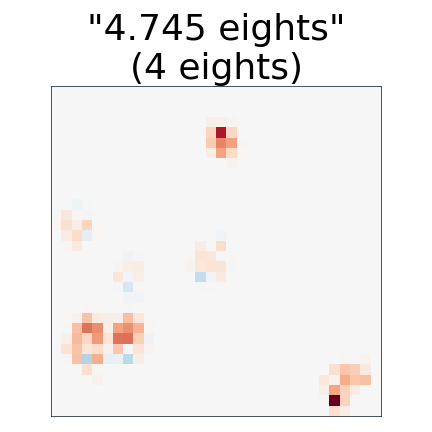}
			   {white.png}
			   {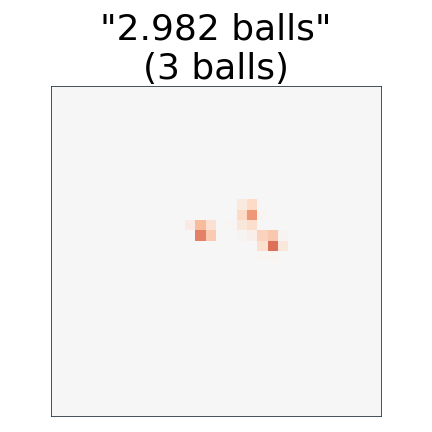}
			   {plotsrd-2_f8c12_resultspngdensity_maps2_2.png}
			   {0}	
    \caption{#1}\label{#2}
\end{figure}

}

\newcommand{\onebytwoimg}[4]{

\begin{figure}
    \centering
    \begin{subfigure}[b]{0.48\textwidth}
        \includegraphics[width=\textwidth]{#1}
        \caption{}
        \label{#4:a}
    \end{subfigure}
    ~ 
    \begin{subfigure}[b]{0.48\textwidth}
        \vspace{-5em}
        \includegraphics[width=\textwidth]{#2}
        \caption{}
        \label{#4:b}
    \end{subfigure}
    \caption{#3}\label{#4}
\end{figure}

}

\newcommand{\mnisttwofeightceightresultsfourspercent}{100.00\ \mathrm{\%}\xspace}

\newcommand{\mnistonefeightceightresultsfivespercent}{100.00\ \mathrm{\%}\xspace}

\newcommand{\mnisttenfeightceightresultsfivespercent}{98.08\ \mathrm{\%}\xspace}

\newcommand{\mnisttwooccfeightceightresultsfourspercent}{84.82\ \mathrm{\%}\xspace}

\newcommand{\mnisttwooccfeightceightresultseightspercent}{88.50\ \mathrm{\%}\xspace}

\newcommand{\mnisttwooccvsfeightceightresultsfourspercent}{56.68\ \mathrm{\%}\xspace}

\newcommand{\mnisttwooccvsfeightceightresultseightspercent}{54.72\ \mathrm{\%}\xspace}

\newcommand{\rdonefeightctwelveresultsduckspercent}{99.39\ \mathrm{\%}\xspace}

\newcommand{\rdtwofeightctwelveresultsduckspercent}{99.58\ \mathrm{\%}\xspace}

\newcommand{\rdtwofeightctwelveresultsballspercent}{99.38\ \mathrm{\%}\xspace}

\title{Weakly-supervised multi-class object localization using only object counts as labels}

\author{%
  Kyle Mills\\
  Faculty of Science\\
  University of Ontario Institute of Technology\\
  Oshawa, Ontario, Canada\\
  Vector Institute for Artificial Intelligence,\\
  Toronto, Ontario, Canada\\
  \texttt{kyle.mills@uoit.net} \\
   \And
   Isaac Tamblyn \\
   National Research Council Canada \\
   Ottawa, Ontario, Canada \\
  Vector Institute for Artificial Intelligence,\\
  Toronto, Ontario, Canada\\
   \texttt{isaac.tamblyn@nrc.ca} \\
}

\begin{document}

\maketitle

\begin{abstract}
We demonstrate the use of an extensive deep neural network to localize instances of objects in images. The EDNN is naturally able to accurately perform multi-class counting using only ground truth count values as labels.  Without providing any conceptual information, object annotations, or pixel segmentation information, the neural network is able to formulate its own conceptual representation of the items in the image.  Using images labelled with only the counts of the objects present, the structure of the extensive deep neural network can be exploited to perform localization of the objects within the visual field. We demonstrate that a trained EDNN can be used to count objects in images much larger than those on which it was trained.  In order to demonstrate our technique, we introduce seven new datasets: five progressively harder MNIST digit-counting data sets, and two data sets of 3d-rendered rubber ducks in various situations. On most of these datasets, the EDNN achieves greater than 99\% test set accuracy in counting objects. 
\end{abstract}

\section{Introduction}

The goal of automated object localization research is to take a two-dimensional projection of a scene (e.g. a photograph) and construct a spatial density map, indicating where in the image the objects of interest appear. The integral of this density map can be evaluated to arrive at a count of the number of objects of various types present in the full three-dimensional area \cite{divideandcount}. There are numerous applications where accurate counting of objects from a visual camera signal is beneficial, such as population monitoring in the Seregeti\cite{snapshotserengeti}, counting humans in crowded locations \cite{Han2017,Hu2016a,Zeng2018a,singleimagecrowdcounting,crossscenecrowdcounting,Privacy-preserving-crowd-monitoring}, or counting bacteria on microscope slides \cite{countception}.  
In order to train supervised computer vision and machine learning methods for this task, labels are required. In the most strongly-supervised techniques, each pixel is assigned to a class and the task is called segmentation, with the neural network tasked at predicting labels at pixel resolution.  Coming in slightly weaker are datasets labelled with bounding boxes around objects and the problem is cast as one of detection.  More weakly-supervised techniques that use point annotations have been employed to count cells \cite{learningtocountwithregression,learningtocountobjectsinimages,cnnboosting,microscopycellcountingcnn,cellcountingbyregressionxue,countception,countingcellsrnntimelapse,ppp_dcc}, cars \cite{imagebasedcnntraffic,overlappingvehiclecounting,countingcarsperspectivefree}, and penguins \cite{penguins,ppp_dcc}, among other things.  Point annotations are more appropriate when object occlusion is present and bounding boxes or pixel-specific labels would overlap significantly \cite{microscopycellcountingcnn}. 
The main difficulty with these approaches lies in obtaining the detailed annotation labels (e.g. bounding boxes or positions). These annotations are traditionally provided by humans, such as the task of clicking on penguins in the arctic \cite{penguins}.  Crowdsourcing this tedious work has become common, but with non-experts and anonymous users providing the labels, a training set could easily become fouled with erroneous labels, making training an accurate model difficult \cite{Sindagi2018}. It is substantially easier to arrive at a count of the number of objects present in the field of view than it is to assign all pixels to class or draw bounding boxes around objects.

\subsection{Related work}

Patch-based counting and localization approaches are common, with the standard approach being to cast the problem as a density-map regression problem. For example, Refs. \cite{learningtocountwithregression,countingcarsperspectivefree, overlappingvehiclecounting,countception} blur point annotations to construct a density map. Then they train a regression model to, acting on a single patch at a time, reproduce this density map. Once the model is trained, a density map can be constructed from new images, providing localization information, and the integral of the density map can provide total count information.

Our application aims to make the task even simpler; knowing only the number of objects present in an image and using this as a label, we train an end-to-end deep learning model capable of achieving both accurate multi-class counting and multi-class localization of objects.  One of the benefits of using only the raw object counts is that count information is more easily obtained than precise location annotations, and furthermore, the count signal need not originate from the visual signal itself.  Measurements from other devices, e.g. weight sensors, optical tripwires, etc. can be used as labels to train a model operating on the visual signal. Unlike other similar approaches \cite{fullyconvolutional}, this approach relies only on a single scalar label, and ground truth segmentation data is not required for obtaining object localization or counting. 
Once trained, the model can be applied to arbitrarily large images. Because the neural network only acts on a small subset of the input image, the actual convolutional neural network model can be relatively small, and the systematic way in dividing input images into evenly-sized patches is very easy to implement.

\section{Methods}
Extensive deep neural networks (EDNN) were proposed by Mills, et. al \cite{Mills2019} as a way to capture the extensive nature of physical properties in physics and chemistry applications. An extensive property is one that scales linearly with system size (such as number of objects, or total energy in the case of chemistry applications), with the obverse being an intensive quantity (such as temperature or density).
The EDNN technique itself is relatively simple: break the input image $x\in\mathbb{R}^{W\times H \times d}$ (where $d$ is the number of channels, e.g. 1 for grayscale and 3 for RGB) into non-overlapping patches of size $f\times f\times d$, called focus regions.
These focus regions are then padded with a border of width $c$, adding ``context'' around the non-overlapping patches.  Each $(f+2c)\times(f+2c)\times d$ ``tile'' is then fed through the same neural network, outputting a single number for each tile.
These results are summed and the final summed value is used as the ``prediction'' against which to compute the loss and perform back-propagation.
The effect of this technique is that the neural network only needs to learn to predict a fractional contribution of the final object count.
The neural network learns automatically how to treat the overlapping context regions so as not to double count contributions. 
Furthermore, since each tile is comprised of non-overlapping focus regions, one can consider the neural network output for a single tile the fractional count of the number of objects present and a density map can be constructed over the original image at a resolution determined by the focus size $f$. 
In the case of small focus, this can enable the precise localization of objects in the image.

\DatasetExamples{An example image from each of the 7 datasets presented in this paper.  The datasets are described in Table \ref{dataset_summary} and additional examples are included in the Supplementary Information.}{ex}

\begin{table}
\begin{tabularx}{\linewidth}{l|X|l }
  Name & Description &  $L_\mathrm{max}$  \\
  \hline
  
  (a) MNIST-1 
  & Single category collage of hand-drawn examples of the digit 5 from the original MNIST data set \cite{Lecun1998a}.  No digits overlap and all are the same size.
  & 25
  \\\hline

  (b) MNIST-2
  & Two category collage of hand-drawn examples of the digit 4 and 8 from the original MNIST data set.  No digits overlap and all are the same size.
  & 12
  \\\hline

  (c) MNIST-10
  & Ten category collage of hand drawn digits from the original MNIST data set.  No digits overlap and all are the same size
  & 6
  \\\hline

  (d) MNIST-2-occ
  & Two category collage of hand drawn examples of the digit 4 and 8 from the original MNIST data set. Digits are permitted to overlap (\textbf{occ}lude) other digits. All digits are the same size. 
  & 15
  \\\hline

  (e) MNIST-2-occ-vs
  & Two category collage of hand drawn examples of the digit 4 and 8 from the original MNIST data set. Digits are permitted to overlap (\textbf{occ}lude) other digits, and the digits are scaled randomly before placement (using a standard scaling function employing bicubic interpolation).
  & 15
  \\\hline
  
  (f) RD-1
  & Single category 3d renderings of zero to five rubber ducks in different scenes.  The ducks are scaled and placed randomly and partial occlusion is permitted (full occlusion is prevented). Each image is $256\times 256$ pixels with three (RGB) channels.
  & 5
  \\\hline

  (g) RD-2
  & Two category 3d renderings of zero to five rubber ducks, and zero to five floating spheres (``balls'').  Some high-number count combinations are omitted (e.g. 5 ducks and 5 balls) due to the difficulty of packing the objects while preventing occlusion.  The spheres are randomly coloured and scaled and randomly assigned either a matte or metallic finish.  Each image is $256\times 256$ pixels with three (RGB) channels.
  & 5
  \\\hline\\

\end{tabularx}
\caption{Summary of the designed datasets. $L\sub{max}$ denotes the largest label value (i.e. the maximum count of each object in a given image).  \label{dataset_summary}}
\end{table}

We have designed multiple data sets through which we demonstrate the EDNN-counting and localization.  The MNIST variants are constructed by extracting 4800 examples of each of the ten numerals from the original MNIST dataset. For each of the ten digits, 480 members are reserved for the testing sets and 4320 are used to construct the training sets (i.e. no unique MNIST example is present in both the training and testing sets) The examples are constructed by choosing a random number $N_i$ between 0 and $L_\mathrm{max}$ for each numeral $i$ present in the data set.  This will serve as the label. $N_i$ digits are then chosen randomly from either the testing or training subset and are composited randomly on an empty $256\times 256$ pixel image through element-wise summation.  In the case where occlusion is permitted, pixel values are clipped at 255. The images are saved as greyscale 8-bit PNG images, with labels in a separate JSON file.

The RD-2 dataset is generated using a script inspired by the CLEVR \cite{clevr} dataset, using Blender to perform the 3d rendering.  For each of the label-scene pairs (e.g. ``5 ducks, 2 balls, mountain scene''), we generated 1024 images by randomly placing, scaling, and rotating the appropriate number of ducks and balls. 896 images of each scene-label pair are used for training and 128 for testing.  The RD-1 dataset is a subset of the RD-2 dataset (the examples with zero balls).  The images are generated at a resolution of $256\times 256$ pixels and stored in RGB (3 channel) 8-bit PNG images. Labels are stored in a separate JSON file.

An important consideration in the design of an EDNN was how to choose an appropriate tile size (e.g. focus and context region).  Mills et. al. discusses this process based on the length scale of the physics of the underlying problem.  In the case of counting variable-sized objects present in an image, the process for choosing a focus and context is less clear. From intuition, we can suggest that the total tile size ($f+2c$) should be large enough to capture an identifiable feature of the largest objects we wish to count.  For example, to count camera-facing waterfowl, one does not need to actually be able to identify a bird; an accurate count could be obtained by only learning to identify beaks.  We found that using a focus of $f=8$ and a context of $c=8$ worked well for the MNIST counting purposes. For the rubber duck counting, we used a slightly larger context, $c=12$, as the largest ducks covered more pixels than the MNIST digits. We will discuss the benefits of different focus and context values further in the discussion.

With our choice of focus and context, the neural network must be designed to act on tiles of size $(f+2c)\times(f+2c)$, $24\times 24$ for MNIST and $32\times 32$ for the ducks.  The input images are zero-padded with $c$ pixels on all sides, and the tiles are constructed using standard TensorFlow \cite{GoogleResearch2015} image functions operating on these zero-padded copies of the input data. 

We used a standard convolutional neural network with $N = \mathrm{floor}(\log2(f+2c)-1)$ layers operating with $K=64$ square kernels of size $k=4$.  Each layer operated with stride $S=2$.  The output of the final convolutional network is flattened and passed into a dense layer with 1024 outputs, which is then fed into a final dense layer. The final layer has $l$ outputs, where $l$ is the dimensionality of the labels (i.e. the number of classes being counted).

In practice, the computational graph is constructed to take a batch of $N\sub{batch}$ $L\times L$ images, and deconstruct them into $N\sub{tiles}=L^2 / f^2$ tiles. The input data is concatenated along the first axis; this results in a tensor of shape $[N\sub{batch}\times N\sub{tiles}, f+2c, f+2c]$. The convolutional neural network operates on this tensor, performing the same operations on all tiles of all images in the batch, reducing it to shape $[N\sub{batch}\times N\sub{tiles}, l]$.  This tensor is then reshaped to $[N\sub{batch}, N\sub{tiles}, l]$; in practice, this can be thought of ``how much stuff'' should be attributed to the $f\times f$ focus region.  We will refer later to this tensor, so we will label it $\mathcal{C}$. Then for the core of the EDNN technique: a summation reduction over the second axis. This results in a $[N\sub{batch}, l]$ tensor denoting the prediction of the $l$ extensive quantities for each image in the batch.  The loss is computed as the mean squared error between this vector and the vector of labels, and the Adam optimizer \cite{Kingma2014} is used to minimize this loss. 

Standard neural network training procedures were employed using TensorFlow \cite{GoogleResearch2015} and the Adam optimizer \cite{Kingma2014} with a learning rate of $10^{-4}$. We trained the EDNN until the loss dropped below $10^{-3}$, between 100 and 500 epochs, depending on the difficulty of the dataset.

\section{Results}

\subsection{MNIST variants}

The simplest case is the MNIST-1 dataset.  The EDNN is able to count the number of fives with $\mnistonefeightceightresultsfivespercent$ accuracy.  Localization of the digits within the input image can be achieved by assigning the contents of the tile contribution tensor $\mathcal{C}$ to the focus regions over which its contributions were obtained.  Doing so results in the ability to detect not only the number of objects in the visual field, but additionally pinpoint their location with considerable accuracy. 


Next is the MNIST-2 dataset, testing the ability of the EDNN to perform mutli-class counting and localization.  The EDNN is tasked with counting 4s and 8s, and does so with $\mnisttwofeightceightresultsfourspercent$
accuracy for both categories.
 Similarly, localization can be achieved.  Interestingly, the EDNN learns to assign a positive contribution to the class of interest, while assigning a negative contribution to the other "negative" class.  This is because of the EDNN's limited receptive field; since it only sees a small region of the input image, in the tiles surrounding a digit, the EDNN cannot see enough information to identify which class the digit belongs to, and thus assigns a small, positive contribution.  Then when the EDNN sees a tile more central to the negative-class digit, it must output a negative value to compensate for its misclassification of the exterior regions.  The result is still a very precise localization of each class of objects in the receptive field.


The MNIST-10 dataset includes examples of all ten handwritten digits, from 0 through 9. The EDNN performs exceptionally well; it performed worst at counting fives, but still performed with $\mnisttenfeightceightresultsfivespercent$ accuracy. The performance of the EDNN on MNIST-10 is shown in \figref{Results}.


A more difficult challenge is when the digits are permitted to overlap (occlude) other digits. Nonetheless, the EDNN is able to count the digits quite accurately with an accuracy of $\mnisttwooccfeightceightresultsfourspercent$ and $\mnisttwooccfeightceightresultseightspercent$ for counting fours and eights, respectively. 


When variable-sized digits are included in the dataset, and digits are allowed to occlude each other, the task is considerably more difficult with only $\mnisttwooccvsfeightceightresultsfourspercent$ and $\mnisttwooccvsfeightceightresultseightspercent$ accuracy for counting fours and eights, respectively.  This is not surprising; looking at the example image in \figref{Results}, it is difficult to differentiate many of the digits even by eye.


The error distributions for all datasets are shown in \figref{errordistributionall}.

\subsection{Rubber ducks}
Next we move on to the rubber ducks.  The EDNN is able to count variable-sized rubber ducks that are permitted to partially occlude each other, and does so with high accuracy ($\rdonefeightctwelveresultsduckspercent$).  The dataset includes five different scenes and a variety of camera angles and lighting conditions affording the EDNN the opportunity to learn to identify objects and not merely base its prediction on the presence of a particular colour in the image. 

Next we trained an EDNN from scratch on the RD-2 dataset, including multi-colored balls in addition to the rubber ducks.  The EDNN performs exceptionally well, counting the two distinct object classes, ``ducks'' and ``balls'', with $\rdtwofeightctwelveresultsduckspercent$ and $\rdtwofeightctwelveresultsballspercent$ accuracy, respectively on the test images. An example is shown in \figref{Results}.

\onebytwoimg{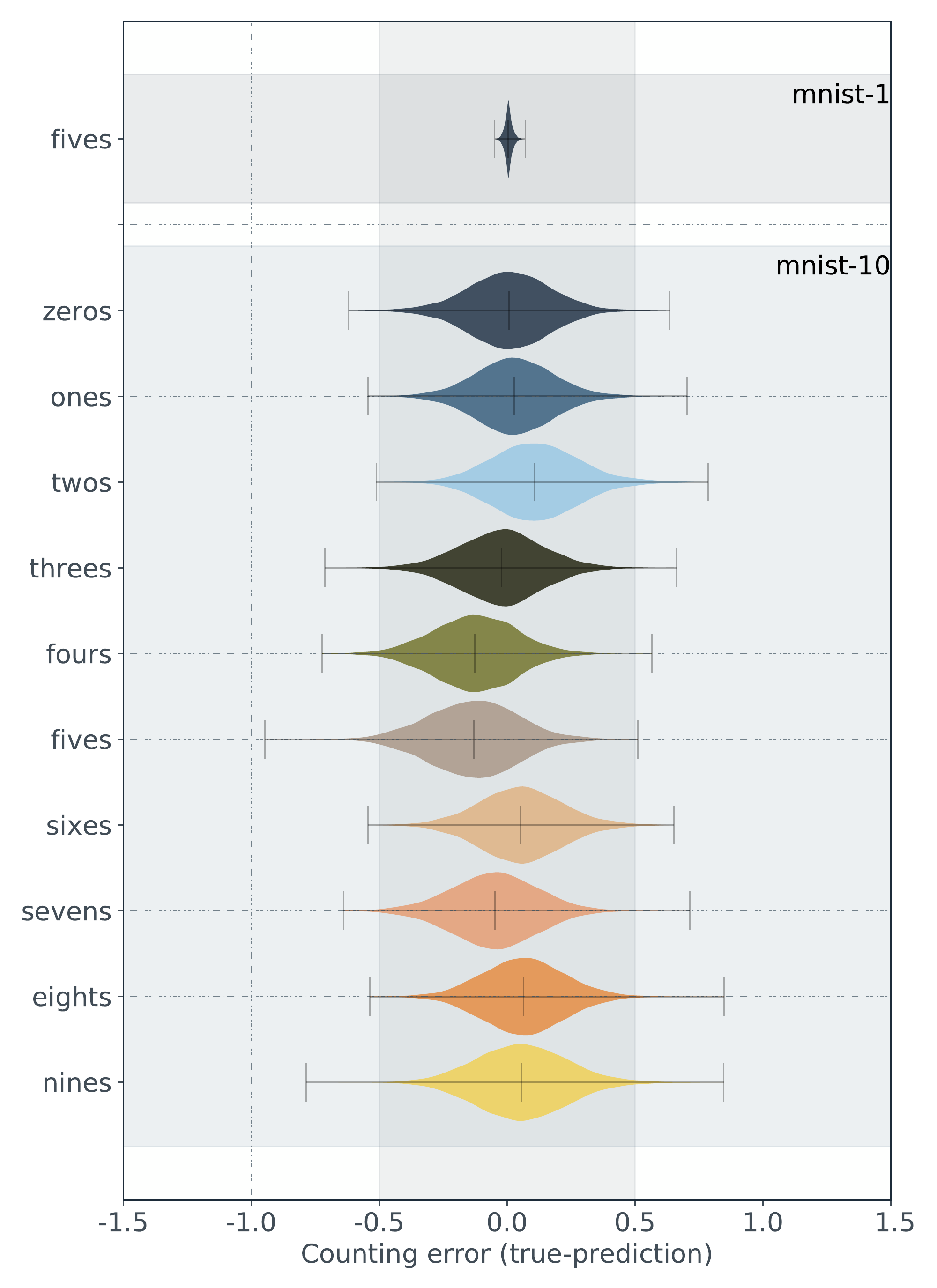}
            {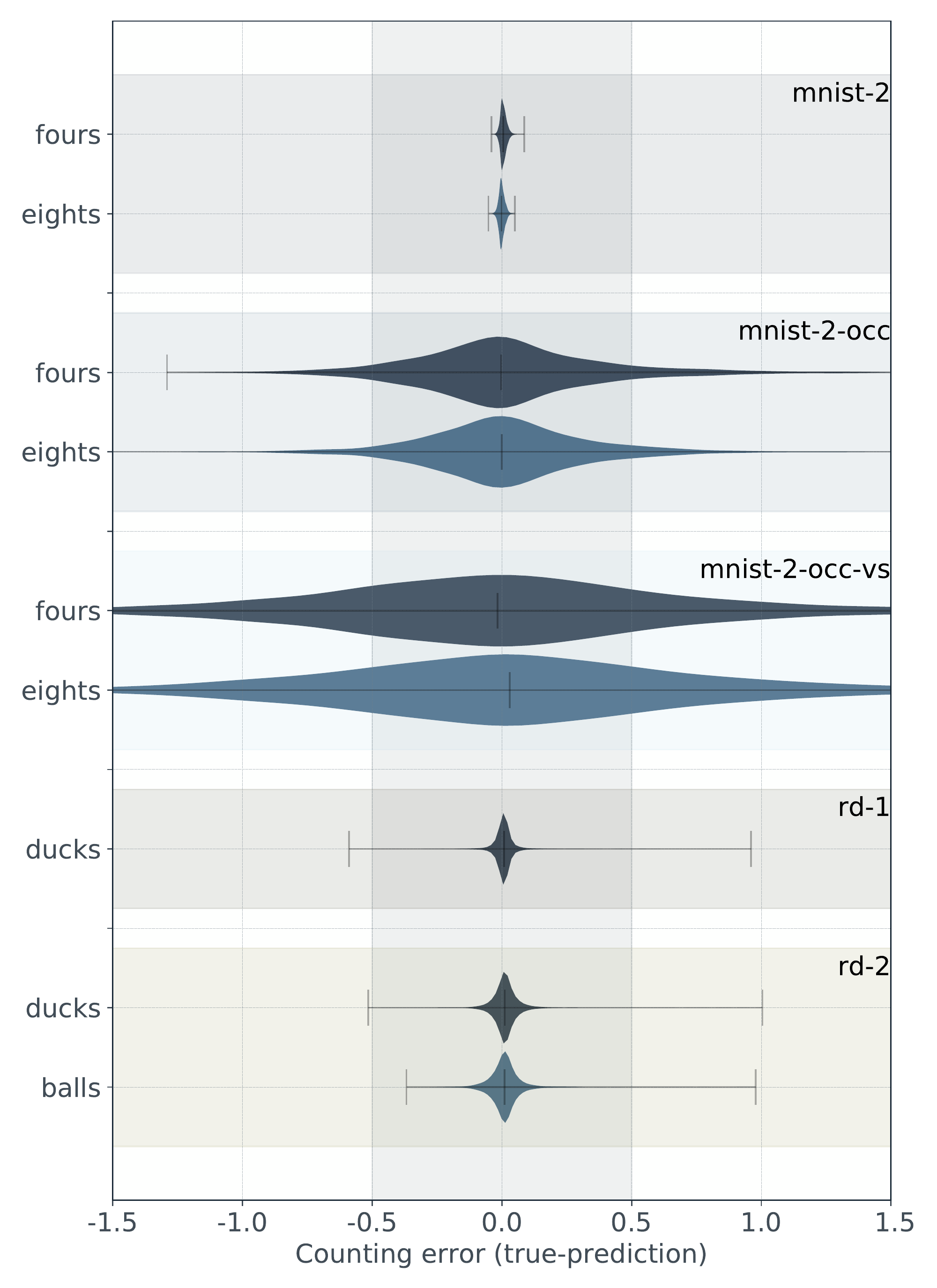}
            {Error distributions for all classes presented in this paper. The minimum, median, and maximum error for each class is displayed as vertical lines on the distributions.  The interval of correct counting ($-0.5$ to $0.5$) is shaded. A rounding of the final count to integer values would result in any examples within this interval being counted correctly.  We can see easily from these plots that the MNIST digit counting with occlusion is the most difficult task.}
            {errordistributionall}



\ResultExamples{Output of the EDNN for counting and localization of a testing example from three datasets: MNIST-10, MNIST-2-occ-vs, and RD-2. The top row represents the reference image $x$. The second and third rows show the tile contributions $\mathcal{C}$ for two of the multi-class labels, e.g. $l_0, l_N$. Red denotes a large positive contribution, whereas blue denotes a negative contribution.  Above, the predicted object count is displayed (the integral (sum) of $\mathcal{C}$) alongside the true count in parentheses.}{Results}

\begin{figure}
\centering
\includegraphics[width=\textwidth]{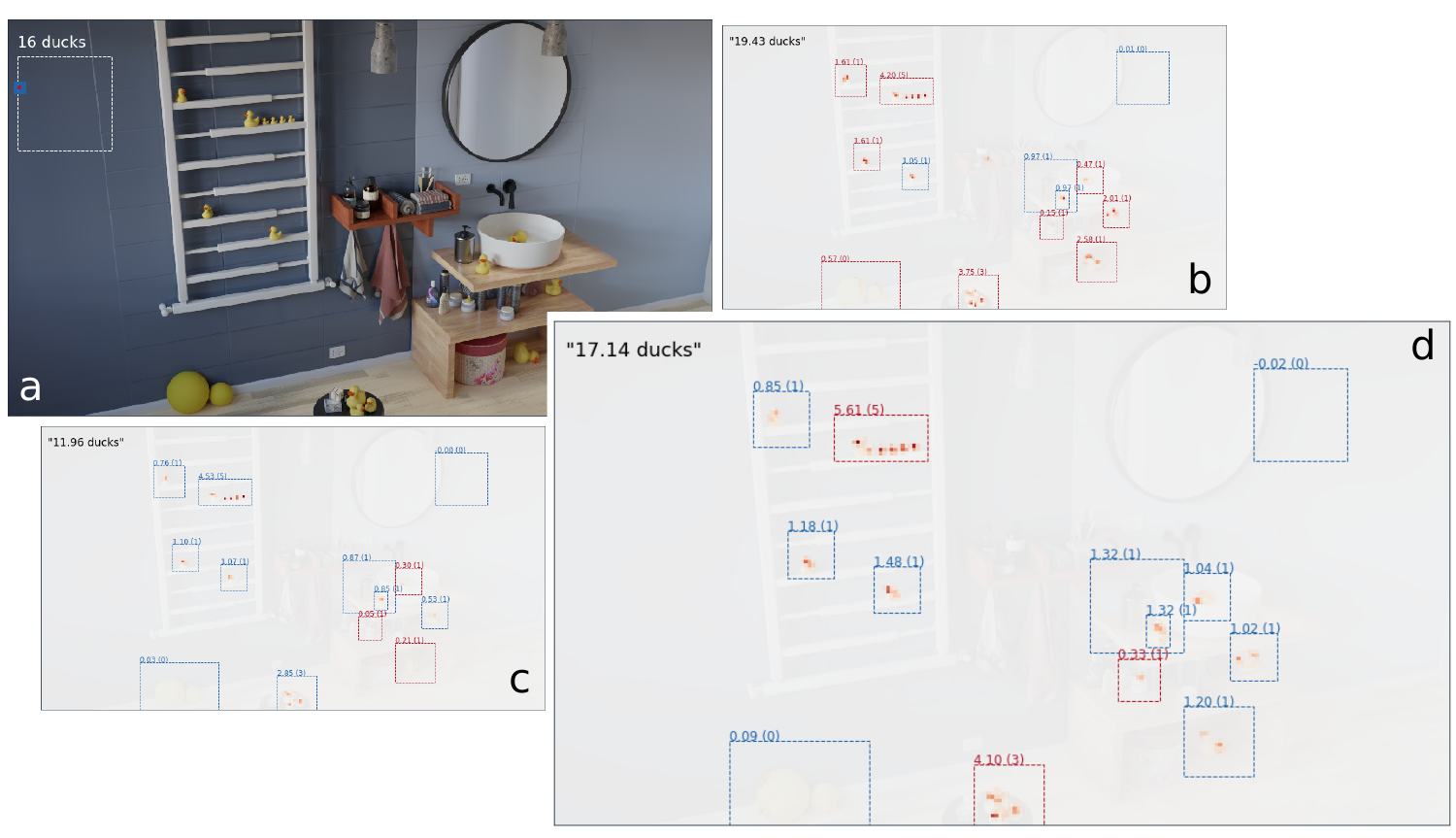}

\caption{We used the models trained on the RD-1 and RD-2 datasets to count the number of ducks in a completely new scene (``bathroom''), at a resolution of $1920\times 1080$.  We show the scene in (a), which contains 16 rubber ducks of various sizes.  The dashed-line inset in (a) denotes the size of the training dataset images, $256\times 256$ pixels, and a tile is displayed, showing both the focus (red) and context (blue) regions to scale.  In (b) and (c) we overlay the contents of the $\mathcal{C}$ tensor, with some regions highlighted with dashed rectangles.  The sums of the interior regions bounded by the rectangles are printed above, as well as the true number of ducks contained within each region. Regions containing a correct count (post-rounding) are shown in blue, whereas regions with an incorrect count are shown in red.  In (b), the RD-1 model performs quite poorly, misidentifying the yellow balls as parts of a duck, and miscounts many regions.  In (c), the RD-2 model performs well, counting most groups of ducks quite accurately.  It correctly ignores the yellow balls, as during training it has seen yellow balls.  Three ducks are miscounted; however, these ducks are either in an orientation or of a size not present in the training set, and thus misidentification is expected. In (d), we fix this by further training the model on an augmented pipeline with rotations of the original images included.  This results in better overall results, including the fallen ducks and identifying the tiny duck.\label{bathroom-results}}

\end{figure}

One might question whether the trained model generalizes to rubber ducks which find themselves in situations not present in the training set.  To test this, we construct a 3d-scene of multiple ($l_0=16$) ducks in a bathroom scene \cite{Tirindelli2018}.  The ducks are various sizes, in various orientations and surrounded by other 3d objects.  The overall image is rendered at a much higher resolution ($1920\times1080$) than the images present in the training set, however the EDNN technique can easily handle this increase in size; there are simply more tiles to evaluate prior to the final summation.  In fact, EDNN can be evaluated on arbitrary input sizes so long as the image resolution is a multiple of the focus region (zero padding can be employed if this is not the case, effectively making this constraint moot).  Most of the ducks in the larger image are of comparable scale to those in the training set.  \figref{bathroom-results}a shows the scene.  The EDNN trained on the duck-only RD-1 dataset does poorly on evaluating the number of ducks in the bathroom.  By looking at $\mathcal{C}$ we can form a hypothesis as to why. Summing the interior of a region of $\mathcal{C}$ will tell us how many ducks the EDNN has decided are present within the boundary of the region.  When we do this for some regions-of-interest, we can see that many ducks are over-counted, while others are under-counted.  It is not possible to tell exactly what the problem is, however the EDNN is clearly considering the yellow balls to be somewhat duck-like.  Our best guess for this failure is that the EDNN is placing too much reliance on the boundary between ``yellow'' and ``non-yellow'' as a good indicator of a duck, and is thus miscounting both large ducks and yellow balls.  If this is indeed the case, the model trained on RD-2 should perform better as the training examples included yellow balls that the EDNN would have needed to learn are not ducks.

\figref{bathroom-results}c shows an identical analysis for the model trained on the RD-2 dataset evaluated on the bathroom scene.  The model misses three ducks, but this time the reason is clear: the erroneous counts are ducks that differ from the ducks in the training set.  One duck is significantly smaller than any ducks present in the training set, and the other two have fallen over, an orientation absent from the training set.

A remedy for such an issue should be clear; train the EDNN additionally on images of sideways rubber ducks.  This can be accomplished most simply by applying a random rotation by an integer multiple of $\pi / 2$ to the input image pipeline during training, augmenting the data set. After doing this, the EDNN does indeed count the sideways rubber ducks and additionally identifies the tiny duck, although at a reduced count since it is smaller. Variations in object sizes can also be handled through data augmentation, scaling down the input images and zero-padding the boundary.

\section{Conclusion}
We demonstrate the use of an extensive deep neural network for providing multi-class object localization in images, using weakly-supervised learning. By training an EDNN to count objects in images, we arrive at a model which can both accurately count instances of objects in images as well as spatially localize them using only object counts as labels.  We demonstrate that multiple classes of objects can be counted simultaneously (multi-class counting). The EDNN is, through training, able to develop an internal representation of the objects suitable for counting without any bounding boxes, point annotations, or ground truth segmentation data. The spatial structure of the EDNN can be exploited to additionally provide a density map of the objects in the image, providing accurate localization within the field of view (multi-class localization). Once trained, we demonstrate the EDNN can be used on images significantly larger and different than those present in the training dataset.  The EDNN technique is a simple and useful technique for computer vision applications.

\bibliographystyle{unsrt}
\bibliography{main}

\begin{thebibliography}{10}

\bibitem{divideandcount}
Tobias Stahl, Silvia~L Pintea, and Jan~C. van Gemert.
\newblock {Divide and Count: Generic Object Counting by Image Divisions}.
\newblock {\em IEEE Transactions on Image Processing}, 28(2):1035--1044, feb
  2019.

\bibitem{snapshotserengeti}
Alexandra Swanson, Margaret Kosmala, Chris Lintott, Robert Simpson, Arfon
  Smith, and Craig Packer.
\newblock {Snapshot Serengeti, high-frequency annotated camera trap images of
  40 mammalian species in an African savanna}.
\newblock {\em Scientific Data}, 2:150026, jun 2015.

\bibitem{Han2017}
Kang Han, Wanggen Wan, Haiyan Yao, and Li~Hou.
\newblock {Image Crowd Counting Using Convolutional Neural Network and Markov
  Random Field}.
\newblock pages 1--6, jun 2017.

\bibitem{Hu2016a}
Yaocong Hu, Huan Chang, Fudong Nian, Yan Wang, and Teng Li.
\newblock {Dense crowd counting from still images with convolutional neural
  networks}.
\newblock {\em Journal of Visual Communication and Image Representation},
  38:530--539, 2016.

\bibitem{Zeng2018a}
Lingke Zeng, Xiangmin Xu, Bolun Cai, Suo Qiu, and Tong Zhang.
\newblock {Multi-scale convolutional neural networks for crowd counting}.
\newblock {\em Proceedings - International Conference on Image Processing,
  ICIP}, 2017-Septe:465--469, 2018.

\bibitem{singleimagecrowdcounting}
Yingying Zhang, Desen Zhou, Siqin Chen, Shenghua Gao, and Yi~Ma.
\newblock {Single-Image Crowd Counting via Multi-Column Convolutional Neural
  Network}.
\newblock In {\em 2016 IEEE Conference on Computer Vision and Pattern
  Recognition (CVPR)}, pages 589--597. IEEE, jun 2016.

\bibitem{crossscenecrowdcounting}
{Cong Zhang}, {Hongsheng Li}, Xiaogang Wang, and {Xiaokang Yang}.
\newblock {Cross-scene crowd counting via deep convolutional neural networks}.
\newblock In {\em 2015 IEEE Conference on Computer Vision and Pattern
  Recognition (CVPR)}, pages 833--841. IEEE, jun 2015.

\bibitem{Privacy-preserving-crowd-monitoring}
Antoni~B. Chan, {Zhang-Sheng John Liang}, and Nuno Vasconcelos.
\newblock {Privacy preserving crowd monitoring: Counting people without people
  models or tracking}.
\newblock In {\em 2008 IEEE Conference on Computer Vision and Pattern
  Recognition}, pages 1--7. IEEE, jun 2008.

\bibitem{countception}
Joseph~Paul Cohen, Genevieve Boucher, Craig~A. Glastonbury, Henry~Z. Lo, and
  Yoshua Bengio.
\newblock {Count-ception: Counting by Fully Convolutional Redundant Counting}.
\newblock mar 2017.

\bibitem{learningtocountwithregression}
{Luca Fiaschi ; Ullrich Koethe ; Rahul Nair ; Fred A. Hamprecht}.
\newblock {Learning to Count with Regression Forest and Structured Labels}.
\newblock In {\em Proceedings of the 21st International Conference on Pattern
  Recognition (ICPR2012)}. [IEEE], 2012.

\bibitem{learningtocountobjectsinimages}
Antoni~B. Chan, {Zhang-Sheng John Liang}, and Nuno Vasconcelos.
\newblock {Privacy preserving crowd monitoring: Counting people without people
  models or tracking}.
\newblock In {\em 2008 IEEE Conference on Computer Vision and Pattern
  Recognition}, pages 1--7. IEEE, jun 2008.

\bibitem{cnnboosting}
Elad Walach and Lior Wolf.
\newblock {Learning to Count with CNN Boosting}.
\newblock pages 660--676. Springer, Cham, 2016.

\bibitem{microscopycellcountingcnn}
Weidi Xie, J.~Alison Noble, and Andrew Zisserman.
\newblock {Microscopy cell counting and detection with fully convolutional
  regression networks}.
\newblock {\em Computer Methods in Biomechanics and Biomedical Engineering:
  Imaging {\&} Visualization}, 6(3):283--292, may 2018.

\bibitem{cellcountingbyregressionxue}
Yao Xue, Nilanjan Ray, Judith Hugh, and Gilbert Bigras.
\newblock {Cell Counting by Regression Using Convolutional Neural Network}.
\newblock pages 274--290. Springer, Cham, 2016.

\bibitem{countingcellsrnntimelapse}
Alexander~Gomez Villa, Augusto Salazar, and Igor Stefanini.
\newblock {Counting Cells in Time-Lapse Microscopy using Deep Neural Networks}.
\newblock jan 2018.

\bibitem{ppp_dcc}
Mark Marsden, Kevin Mcguinness, Suzanne Little, Ciara~E Keogh, and Noel~E
  O'connor.
\newblock {People, Penguins and Petri Dishes: Adapting Object Counting Models
  To New Visual Domains And Object Types Without Forgetting}.
\newblock Technical report, 2017.

\bibitem{imagebasedcnntraffic}
Jiyong Chung and Keemin Sohn.
\newblock {Image-Based Learning to Measure Traffic Density Using a Deep
  Convolutional Neural Network}.
\newblock {\em IEEE Transactions on Intelligent Transportation Systems},
  19(5):1670--1675, may 2018.

\bibitem{overlappingvehiclecounting}
Shiv Surya and Venkatesh~Babu R.
\newblock {TraCount: a deep convolutional neural network for highly overlapping
  vehicle counting}.
\newblock In {\em Proceedings of the Tenth Indian Conference on Computer
  Vision, Graphics and Image Processing - ICVGIP '16}, pages 1--6, New York,
  New York, USA, 2016. ACM Press.

\bibitem{countingcarsperspectivefree}
Daniel O{\~{n}}oro-Rubio and Roberto~J L{\'{o}}pez-Sastre.
\newblock {Towards Perspective-Free Object Counting with Deep Learning}.
\newblock pages 615--629. 2016.

\bibitem{penguins}
Carlos Arteta, Victor Lempitsky, and Andrew Zisserman.
\newblock {Counting in the Wild}.
\newblock pages 483--498. Springer, Cham, 2016.

\bibitem{Sindagi2018}
Vishwanath~A. Sindagi and Vishal~M. Patel.
\newblock {A survey of recent advances in CNN-based single image crowd counting
  and density estimation}.
\newblock {\em Pattern Recognition Letters}, 107:3--16, 2018.

\bibitem{fullyconvolutional}
Jonathan Long, Evan Shelhamer, and Trevor Darrell.
\newblock {Fully Convolutional Networks for Semantic Segmentation}.
\newblock Technical report, 2016.

\bibitem{Mills2019}
Kyle Mills, Kevin Ryczko, Iryna Luchak, Adam Domurad, Chris Beeler, and Isaac
  Tamblyn.
\newblock {Extensive deep neural networks for transferring small scale learning
  to large scale systems}.
\newblock {\em Chemical Science}, 10(15):4129--4140, aug 2019.

\bibitem{Lecun1998a}
Y.~Lecun, L.~Bottou, Y.~Bengio, and P.~Haffner.
\newblock {Gradient-based learning applied to document recognition}.
\newblock {\em Proceedings of the IEEE}, 86(11):2278--2324, 1998.

\bibitem{clevr}
Justin Johnson, Bharath Hariharan, Laurens van~der Maaten, Li~Fei-Fei,
  C.~Lawrence Zitnick, and Ross Girshick.
\newblock {CLEVR: A Diagnostic Dataset for Compositional Language and
  Elementary Visual Reasoning}.
\newblock dec 2016.

\bibitem{GoogleResearch2015}
Mart{\'{i}}n Abadi, Ashish Agarwal, Paul Barham, Eugene Brevdo, Zhifeng Chen,
  Craig Citro, Greg~S. Corrado, Andy Davis, Jeffrey Dean, Matthieu Devin,
  Sanjay Ghemawat, Ian Goodfellow, Andrew Harp, Geoffrey Irving, Michael Isard,
  Yangqing Jia, Rafal Jozefowicz, Lukasz Kaiser, Manjunath Kudlur, Josh
  Levenberg, Dan Mane, Rajat Monga, Sherry Moore, Derek Murray, Chris Olah,
  Mike Schuster, Jonathon Shlens, Benoit Steiner, Ilya Sutskever, Kunal Talwar,
  Paul Tucker, Vincent Vanhoucke, Vijay Vasudevan, Fernanda Viegas, Oriol
  Vinyals, Pete Warden, Martin Wattenberg, Martin Wicke, Yuan Yu, and Xiaoqiang
  Zheng.
\newblock {TensorFlow: Large-Scale Machine Learning on Heterogeneous
  Distributed Systems}.
\newblock {\em None}, 1(212):19, mar 2016.

\bibitem{Kingma2014}
Diederik~P. Kingma and Jimmy Ba.
\newblock {Adam: A Method for Stochastic Optimization}.
\newblock pages 1--15, dec 2014.

\bibitem{Tirindelli2018}
Davide Tirindelli.
\newblock {Blue bathroom}, 2018.

\end{thebibliography}

\end{document}